\documentclass{article} 
\usepackage{collas2022_conference,times}

\usepackage{amsmath,amsfonts,bm}

\def\eqref#1{equation~\ref{#1}}

\def\1{\bm{1}}

\DeclareMathAlphabet{\mathsfit}{\encodingdefault}{\sfdefault}{m}{sl}
\SetMathAlphabet{\mathsfit}{bold}{\encodingdefault}{\sfdefault}{bx}{n}

\usepackage{graphicx}
\usepackage{amsmath}
\usepackage{amssymb}
\usepackage{multirow}
\usepackage{makecell}
\usepackage{subcaption}
\usepackage{algpseudocode}
\usepackage{algorithm2e}
\usepackage{colortbl}
\usepackage{booktabs}

\DeclareUnicodeCharacter{2212}{-}

\usepackage{hyperref}
\hypersetup{
    colorlinks=true,
    linkcolor=red,
    filecolor=magenta,      
    urlcolor=blue,
    citecolor=purple,
    pdftitle={Overleaf Example},
    pdfpagemode=FullScreen,
    }

\makeatletter
\newcommand{\printfnsymbol}[1]{%
  \textsuperscript{\@fnsymbol{#1}}%
}
\makeatother

\title{Curbing Task Interference using Representation Similarity-Guided Multi-Task Feature Sharing}

\author{Naresh Kumar Gurulingan, Elahe Arani\thanks{Equal advising.}, ~Bahram Zonooz\printfnsymbol{1} \\
  Advanced Research Lab, NavInfo Europe, The Netherlands \\
  \texttt{\{naresh.gurulingan, elahe.arani\}@navinfo.eu, bahram.zonooz@gmail.com} \\
}

\collasfinalcopy 

\begin{document}
\maketitle

\begin{abstract}

Multi-task learning of dense prediction tasks, by sharing both the encoder and decoder, as opposed to sharing only the encoder, provides an attractive front to increase both accuracy and computational efficiency. When the tasks are similar, sharing the decoder serves as an additional inductive bias providing more room for tasks to share complementary information among themselves. However, increased sharing exposes more parameters to task interference which likely hinders both generalization and robustness. Effective ways to curb this interference while exploiting the inductive bias of sharing the decoder remains an open challenge. To address this challenge, we propose Progressive Decoder Fusion (PDF) to progressively combine task decoders based on inter-task representation similarity. We show that this procedure leads to a multi-task network with better generalization to in-distribution and out-of-distribution data and improved robustness to adversarial attacks. Additionally, we observe that the predictions of different tasks of this multi-task network are more consistent with each other. Code is made available at \url{github.com/NeurAI-Lab/ProgressiveDecoderFusion}.
\end{abstract}


\section{Introduction}
Obtaining real-time predictions from neural networks is imperative for time-critical applications such as autonomous driving. These applications also require predictions from multiple tasks to shed light on varied aspects of the input scene. Multi-task networks (MTNs) can elegantly combine these two requirements by jointly predicting multiple tasks while sharing a considerable number of parameters among tasks. On the other hand, training separate single task networks could lead to different task predictions contradicting each other. Moreover, each of these networks has to be individually robust to various forms of adverse inputs such as image corruptions and adversarial attacks. MTNs include an inductive bias in the shared parameter space which encourages tasks to share complementary information to improve predictions. This information sharing also enables tasks to provide consistent predictions while holding the potential to improve robustness \citep{mao2020multitask}.

In an ideal scenario, this inductive bias of MTNs would lead to learning representations that generalize better and are more robust, while providing consistent predictions. In practice, however, task interference hinders the attainment of such representations \citep{Zhao2018AMM}. Task interference is an undesired consequence of the shared parameter space enabling tasks to exchange conflicting information. One approach to reduce task interference is to modify the encoder to include isolated task-specific parameters \citep{Liu2019EndToEndML} enabling tasks to encode conflicting information in these isolated parameters. Nevertheless, conflicts in the shared parameters could still occur. To eliminate task interference, other works rely on using independent subnetworks for tasks \citep{Strezoski_2019_ICCV} or only update task-specific parameters with the task loss \citep{10.1007/978-3-030-58565-5_41}. However, \cite{Strezoski_2019_ICCV} do not share parameters between tasks thereby lacking the associated inductive bias of sharing while \cite{10.1007/978-3-030-58565-5_41} limit sharing of complementary information among tasks in the shared parameters. Alternatively, grouping tasks together based on a notion of similarity could potentially mitigate task interference as the more similar the tasks are the less likely they share conflicting information.

Intuitive notions of similarity between tasks might not necessarily hold in the feature space. Therefore, techniques to determine how tasks relate to each other at different layers of an MTN are required. \cite{fifty2021efficiently} quantify inter-task similarity based on the effect of one task's loss when the other task's gradients are used to update shared weights. They use this similarity to group tasks into different multi-task networks. However, tasks could still require learning similar features in some layers, and in these layers, they can be combined. Determining task-sharing at a layer level has not been explored much in the literature. \cite{BranchedVand} use representation similarity analysis to group tasks in the encoder. They group tasks at all layers based on layerwise similarity between fully trained single task networks. However, the similarity between tasks in subsequent layers is likely to change once they have been combined at a certain layer.

\textit{Therefore, to group tasks at a certain layer, it could be necessary to compare representation similarities in light of how the tasks have been grouped in the previous layer.} We propose an algorithm that only groups tasks at a layer based on a fully trained network with grouping done at the previous layer. Specifically, we propose Progressive Decoder Fusion (PDF) algorithm which alternates between fusing tasks at a decoder stage and retraining the fused network. Essentially, the PDF algorithm revolves around the question:

\textit{How to effectively determine a sharing scheme in the decoder to best curb task interference while benefiting from complementary information sharing?}

To test the strengths of the proposed algorithm, we evaluate the final decoder sharing scheme under different settings such as generalization to in-distribution and out-of-distribution data and robustness to adversarial attacks. We observe that the obtained decoder sharing scheme leads to improvements in generalization, robustness, and inference speed while requiring less number of parameters.
Our contributions include:
\begin{itemize}
    \item Progressive Decoder Fusion (PDF) algorithm which groups tasks in a layer considering how tasks have been grouped in previous layers.
    \item Comprehensive evaluation of the final network obtained with the PDF algorithm to test its robustness to adversarial attacks and generalization to in-distribution and out-of-distribution data.
    \item Exploring the role of different tasks with regards to the consistency between predictions, generalization, and robustness.
\end{itemize}

\section{Related Works}
Progress in multi-task learning (MTL) has come from varied directions including custom architectures (Section \ref{subsec: custom_arch}), task balancing (Section \ref{subsec: balance}) and task grouping (Section \ref{subsec: grouping}). Since the primary goal of this work is to curb task interference in MTL, we probe works in each of these categories and discuss how they address task interference.

\subsection{Custom Architectures}
\label{subsec: custom_arch}

One way to alleviate task interference is to introduce task-specific parameters in the shared encoder. This modification could enable the network to encode task information that might conflict with other tasks in the task-specific parameters. In MTAN \citep{Liu2019EndToEndML}, each task is equipped with its own attention modules at different stages of the encoder. \cite{10.1007/978-3-030-58565-5_41} use task-specific 1$\times$1 convolutions after each convolution in the encoder. Only these 1$\times$1 convolutions are trained with the task gradients to explicitly avoid task interference. \cite{Strezoski_2019_ICCV} propose task-specific routing to create randomly initialized task-specific subnetworks to reduce interference. \cite{sun2020adashare} propose to learn task-specific policies with a sharing objective and a sparsity objective to balance the number of ResNet blocks shared between tasks and task interference. These methods likely reduce conflicts on the shared parameter space. However, they require architecture modifications in the encoder and could require training on ImageNet to initialize weights. 

\subsection{Task balancing}
\label{subsec: balance}

Task interference can be seen as the consequence of conflicting task gradient directions in the shared parameters. Task gradients can be modified such that the disagreement between them is reduced to mitigate task interference. The PCGrad \citep{NEURIPS2020_3fe78a8a} uses cosine similarity to identify different pairs of task gradients with contradicting directions in each shared parameter. In each of these pairs, one of the gradients is projected onto the normal vector of the other to reduce conflict. \cite{NEURIPS2020_16002f7a} reduce the probability of using negative task gradients during backward pass thereby reducing gradient conflict. As gradients are modified, these approaches might lose task-specific information.

Individual task losses can be weighted in different ways to address the variance in loss scales \citep{Liu2019EndToEndML, Kendall2018MultitaskLU, Guo2018DynamicTP, Chen2018GradNormGN}. These methods primarily attempt to avoid certain tasks from dominating gradient updates. However, they can only be viewed as a means to loosely modulate task interference as they affect the extent to which tasks affect the shared parameters. Our approach can be used in tandem with task balancing techniques.

\begin{figure}
  \centering
  \includegraphics[width=1\linewidth]{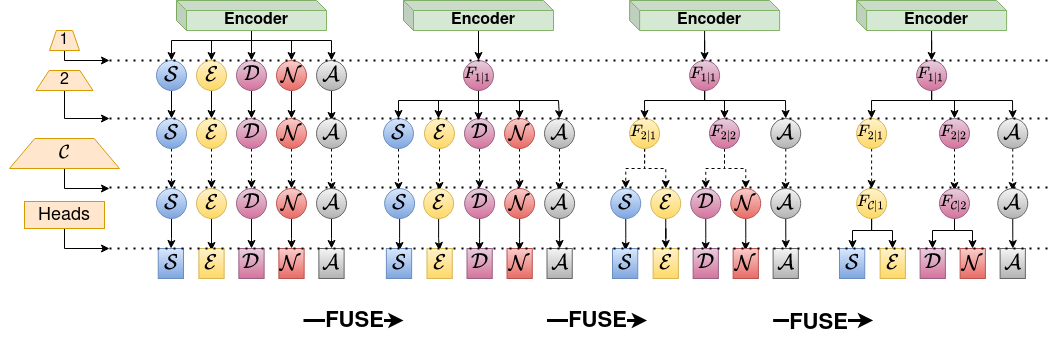}
   \caption{Fusing decoders at different candidate stages. From left to right decoder fusion at candidate stages 1, 2 and $\mathcal{C}$ is shown. $\mathcal{S}, \mathcal{E}, \mathcal{D}, \mathcal{N}$ and $\mathcal{A}$ denote different tasks, namely semantic segmentation, edge detection, depth estimation, surface normals and autoencoder. The dotted horizontal lines show the candidate decoder stages and the task-specific heads. The lines connecting decoder stages in 2 and $\mathcal{C}$ are dotted to show that there are more stages in between. The task-specific heads are not fused in any of the approaches. $\mathcal{F}$ denotes fused decoder and $\mathcal{F}_{i|j}$ denotes $i^{th}$ candidate stage of $j^{th}$ task decoder.}
\label{fig:fusion}
\end{figure}

\subsection{Task grouping}
\label{subsec: grouping}

If only similar tasks are grouped together either at the network level (each task group split into separate networks) or at a layer level (across the network depth), task interference can be reduced. \cite{Standley2020WhichTS} train several MTNs created with all possible combinations of tasks. They pick the combination with the lowest total loss across tasks under a certain computation budget as the desired MTN. \cite{fifty2021efficiently} quantify affinity of task i on task j based on the relative change of task j's loss before and after updating shared parameters with task i. This affinity between different task pairs is accumulated throughout the training and tasks are grouped into different networks such that the overall affinity is maximized.

Other approaches restrict all the tasks to remain in the same network and group them layerwise. Grouping layerwise reduces the required computation but task interference could increase. \cite{pmlr-v119-guo20e} use an automated approach based on Gumbel-Softmax sampling of connections between a child layer and a number of parent layers in a topology. After training, at every child layer, only the connection to the parent layer with the highest probability is retained leading to the effect of tasks being separated at a certain layer. This separation reduces task interference. \cite{BranchedVand} use similarity between representations of trained single task networks to determine how to group tasks at different stages in the encoder. Unlike \cite{BranchedVand}, we study task grouping in the decoder and propose grouping tasks in a progressive fashion.

\section{Sharing task decoders while curbing task interference}

The inductive bias of sharing parameters among different tasks is intuitively desirable as it enables tasks to share complementary information. The early layers learn general features and as we progress deeper through the network, the features learnt become increasingly task-specific. There is often no clear indication as to where in the network the transition between generic to task-specific features happens. In dense prediction tasks, the representations learnt in the decoder for similar tasks might only diverge and become task-specific in the later layers. Most of the early decoder layers could likely be shared among tasks while providing improved generalization, improved robustness, and reduced computational overhead. 

In the upcoming sections, we formally define the problem (Section \ref{subsec: problem}), go over the similarity measure we use (Section \ref{sebsec: similarity}), and discuss our proposed algorithm (Section \ref{sebsec: approach1} to arrive at an effective decoder sharing scheme.

\subsection{Problem Setup}
\label{subsec: problem}

Given a set of $\mathcal{T}$ tasks with each having its own decoder $\mathbb{D}_\mathnormal{t}$, we intend to combine the decoders while incurring limited task interference. All the decoders have the same architecture. We consider $\mathcal{C}$ candidate stages $\{\mathbb{D}_{1|\mathnormal{t}}, \mathbb{D}_{2|\mathnormal{t}},...,\mathbb{D}_{\mathcal{C}|\mathnormal{t}}\}$, $\forall \mathnormal{t} \in \{\mathcal{T}\}$ where the task decoders can be combined. We refer to combining decoders as fusion. For ease of reading, we assign a letter to each task and use this letter to denote the task decoder. The tasks involved and their associated decoders are semantic segmentation: $\mathbb{D}_{1:\mathcal{C}|\mathcal{S}}$, depth estimation: $\mathbb{D}_{1:\mathcal{C}|\mathcal{D}}$, edge detection: $\mathbb{D}_{1:\mathcal{C}|\mathcal{E}}$, surface normals: $\mathbb{D}_{1:\mathcal{C}|\mathcal{N}}$ and autoencoder: $\mathbb{D}_{1:\mathcal{C}|\mathcal{A}}$. Figure \ref{fig:fusion} depicts three different decoder fusions done at candidate stages 1, 2 and $\mathcal{C}$.

\begin{minipage}{.55\linewidth}
\RestyleAlgo{ruled}
\begin{algorithm}[H]
\caption{Progressive Decoder Fusion}
Initialize network with each task having a separate decoder, i.e., $\mathbb{D}_{1:\mathcal{C}|\mathcal{S}} \neq \mathbb{D}_{1:\mathcal{C}|\mathcal{D}} \neq \mathbb{D}_{1:\mathcal{C}|\mathcal{E}} \neq \mathbb{D}_{1:\mathcal{C}|\mathcal{N}} \neq \mathbb{D}_{1:\mathcal{C}|\mathcal{A}}$\;
Task set $\mathcal{T} \gets \{\mathcal{S}, \mathcal{D}, \mathcal{E}, \mathcal{N}, \mathcal{A}\}$\;
Candidate stage $c \gets 0$ (last encoder stage)\;
Train network until convergence\;
\For{$c \in$ \{1,..., $\mathcal{C}$\}}{
  $g \gets$ number of fused groups in stage $c-1$\;
  \For{fused group \{$\mathcal{F}_{c-1|1}, ..., \mathcal{F}_{c-1|g}$\} }{
    $\mathnormal{t} \gets $ number of tasks branching from $g$\;
    \eIf{$\mathnormal{t} = 1$}{
        Continue to next $g$\;
    }
    {
    Measure $\mathnormal{t}\times\mathnormal{t}$ CKA similarity matrix\;
    Run \textit{Grouping Algorithm \ref{algo:1}}\;
    }
  }
  Retrain the fused network until convergence\;
}
\label{algo:2}
\end{algorithm}
\end{minipage}
\hspace{5mm}
\begin{minipage}{.4\linewidth}
\RestyleAlgo{ruled}
\begin{algorithm}[H]
\caption{Grouping}\label{algo:1}
    \textbf{Input:} Task set $\mathnormal{T}$ and $\mathnormal{T}\times\mathnormal{T}$ similarity matrix $S$\;
    Group $\mathnormal{G}$ $\gets$ \{tasks\}\;
    Grouping $\mathcal{G}$ $\gets$ \{groups\}\;
    Task value in group $\mathnormal{G}$ is $\mathcal{V}_t \gets \frac{1}{i} \sum_{i} S_{ti} $, where $i \subset G \backslash {t}$, $S_{ti}$ is the similarity between task t and i\; 
    Group value $\mathcal{V}_\mathnormal{G} \gets \frac{1}{t} \sum_t \mathcal{V}_t$, where t is the \#tasks in group \;
    Unique Grouping $\mathbb{G} \subset \mathcal{G}$, such that all tasks are present exactly once\;
    \For{all unique groupings}{
        $\mathcal{V}_\mathbb{G} \gets \frac{1}{\mathnormal{g}} \sum_g \mathcal{V}_g$, where g is the \#groups in grouping\;
    }
    Final grouping $\gets$ grouping with maximum $\mathcal{V}_\mathbb{G}$\; 
\KwResult{Final grouping}
\end{algorithm}
\end{minipage}


\subsection{Representation similarity}
\label{sebsec: similarity}

In MTL, a well accepted notion is that jointly learning similar tasks would improve overall performance. This notion is intuitively well placed as similar tasks would need similar feature representations which can be attained by sharing parameters instead of using individual networks. However, this notion does not necessarily have to be only at the network level (each task group split into separate networks) but can also be used to combine tasks at the layer level (across the network depth) \citep{pmlr-v119-guo20e, BranchedVand}.

Two tasks can be combined at a particular candidate stage of the MTN if they require learning similar representations at that stage. Central Kernel Alignment (CKA) is a similarity metric with desirable properties such as invariance to orthogonal transformations and isotropic scaling enabling meaningful comparisons of learnt representations \citep{pmlr-v97-kornblith19a}. Further details regarding CKA is provided in the appendix Section \ref{section:CKA_detailed}. At a particular decoder stage, we quantify the pairwise similarity between the representations of all task decoders using CKA. Specifically, given decoder activations $\{\mathbb{D}_{\mathnormal{c}|1}, ..., \mathbb{D}_{\mathnormal{c}|\mathcal{T}}\}$ at candidate stage $\mathnormal{c}$, we construct a pairwise CKA similarity matrix of shape $\mathcal{T}\times\mathcal{T}$ where each element represents similarity between the tasks corresponding to the row and column. Since CKA is symmetric, i.e., CKA($\mathbb{D}_{\mathnormal{c}|i}$, $\mathbb{D}_{\mathnormal{c}|j}$) = CKA($\mathbb{D}_{\mathnormal{c}|j}$, $\mathbb{D}_{\mathnormal{c}|i}$), the $\mathcal{T}\times\mathcal{T}$ similarity matrix is also symmetric. This pairwise similarity matrix is used for task grouping in the PDF algorithm (Section \ref{sebsec: approach1}).

\subsection{Similarity guided Progressive Decoder Fusion}
\label{sebsec: approach1}
In Section \ref{sebsec: similarity}, we saw that CKA can be used to quantify the similarity between two learned representations. Equipped with this, we now look at ways to arrive at a decoder sharing scheme that provides good generalization and robustness.

We first train a network where each of the tasks has its own decoder and calculate the similarity of learned representations at the first candidate stage of the decoder using a subset of training images. With the similarity scores, we group tasks at the first decoder stage using a grouping algorithm. Essentially, this grouping algorithm lists all possible task groupings and identifies a set of groups that cover all tasks exactly once such that the overall similarity is maximized (details in Algorithm \ref{algo:1}).
This new model is reinitialized\footnote{While reinitializing, the parameters are reset to the original intialization where the encoder is initialized with ImageNet pretrained weights while the decoder(s) and task heads are randomly initialized. and retrained after which grouping is done at the second decoder stage among tasks which grouped together in the first stage}. We repeat this process for all decoder stages unless each task branched separately. This approach is outlined in Algorithm \ref{algo:2}.

\begin{figure}
  \begin{minipage}[c]{0.4\textwidth}
    \includegraphics[width=1\linewidth]{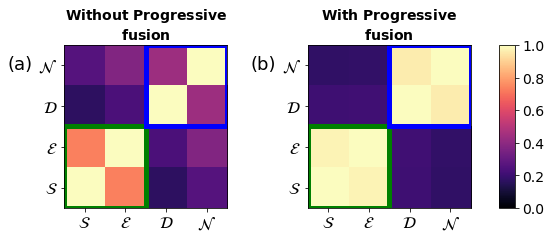}
  \end{minipage}\hfill
  \begin{minipage}[c]{0.55\textwidth}
    \caption{
		Similarity change between tasks at decoder stage 4 without (a) and with progressive fusion (b). Task similarities within both the green and blue boxes differ but within the blue box there is considerable difference.
    } \label{fig:sim_change}
  \end{minipage}
\end{figure}

Unlike \cite{BranchedVand}, we do not group tasks across all decoder stages using the initial network where each task has its own decoder. This difference stems primarily from our experimental observation that learned representations at subsequent candidate stages change once the previous stage is fused as is illustrated in Figure \ref{fig:sim_change}. In (a), the task similarities are determined using the initial network while (b) show task similarities with progressive fusion. We see that the similarities have considerably changed. Thus, fusing decoder parameters at all candidate stages based only on the similarity measure between individual decoders could result in sub-optimal solutions. We present more empirical evidence for change in the similarity between tasks once previous stage decoders are fused in Section \ref{res:prog_motive}.

\section{Role of different tasks}
\label{subsec: consist}

The tasks in an MTN can share complementary information with each other to gain performance improvements due to the inductive bias of shared parameter space. Auxiliary tasks act as a source of complementary information to help improve the performance of main tasks \citep{8578175}. For the task of edge detection, we only consider the delineating edges between regions segmented by semantic segmentation. Essentially, the edge detection task is a binary pixel-wise classification task that predicts the delineating pixels in segmentation ground truth. Thus, semantic segmentation and edge detection are related semantically. On the other hand, depth estimation and surface normals share a geometric relationship \citep{qi2018geonet}. These pairs of tasks could potentially share complementary information and could lead to improved performance overall \citep{8578175}.

Since it is possible to transform semantic segmentation predictions to edge detection and depth estimation predictions to surface normals, we measure two prediction consistencies. We transform semantic segmentation to edge detection and measure semantic consistency as the number of pixels matching between transformed segmentation predictions and edge predictions relative to the total number of pixels (pixel accuracy). Likewise, we measure geometric consistency as the mean of all per pixel cosine similarities between transformed depth predictions and surface normal predictions. The obtained prediction consistencies are presented and discussed in Section \ref{res:consistency}.

The four tasks which we addressed so far provide diverse predictions. This diversity could in and of itself aid in learning a shared representation that exhibits improved generalization as it needs to address these diverse predictions. On top of these tasks, we also include an autoencoder to force the shared representation to not just rely on features that aid in-distribution performance. We then analyze the role of the autoencoder and evaluate whether it aids in generalization and robustness improvements in Section \ref{subsec:ae_role}.

\section{Experiments}
To demonstrate the strengths of the PDF algorithm, we present experimental results obtained with UniNet \citep{Gurulingan_2021_ICCV} architecture. To study fusion in the various stages of the decoder, we pick a hard parameter sharing architecture that shares the entire encoder with all tasks. We consider UniNet which has five different decoder stages branching from the encoder. We use the implementation provided by \cite{fifty2021efficiently} for obtaining the best possible task grouping at a decoder stage using CKA similarity.

We use two datasets for experimentation. \textbf{Cityscapes} \citep{Cordts2016Cityscapes} is a driving scenes dataset consisting of images from various European cities. The dataset consists of 2975 training images and 500 validation images with a resolution of 1024$\times$2048. For training and validation, we use an input resolution of 512$\times$1024. \textbf{NYUv2} \citep{Silberman2012IndoorSA} consists of indoor scene images of resolution 480$\times$640. There are 795 training and 654 validation images. To calculate CKA similarities, we use all the training images and 800 training images for the NYUv2 and Cityscapes datasets, respectively. For both datasets, all numbers are reported on the validation set. CS and NYU denote to the Cityscapes and NYUv2 datasets, respectively.

We refer to complementary information sharing between tasks as \textit{positive transfer}. Also, we collectively refer to inference speed and parameter count of a network as \textit{inference efficiency}. An inference efficient network has high inference speed and low parameter count. We quantify the generalization of all models to in-distribution validation set (IID) and out-of-distribution datasets (OOD) obtained using natural image corruptions \cite{hendrycks2019robustness}. The OOD numbers are reported for four corruption types (\textbf{noise, blur, weather, digital}) as the average across 5 severity levels and all corruptions belonging to the corresponding type. We also quantify the robustness of these models to adversarial attacks with the help of Projected Gradient Descent (PGD) attack conducted using a step size of 1 and for $\min (\epsilon+4,\lceil 1.25 \epsilon\rceil)$ iterations \citep{DBLP:conf/iclr/KurakinGB17a}. PGD numbers are reported for \textbf{low} \pmb{$\epsilon$} (average across $\epsilon$ bounds \{0.25, 0.5, 1\}) and \textbf{high} \pmb{$\epsilon$} (average across $\epsilon$ bounds \{4, 8\}) by using the corresponding task loss as the attack objective. All the numbers are reported as the mean of 3 random seed runs.

We train all the models for 140 epochs using the RAdam optimizer with a learning rate of 0.0001 and stepwise learning rate schedule with a learning rate reduced by a factor of 10 at epochs 98 and 126. The batch size used is 8. The same hyperparameters are used for both datasets. We consider two baselines, one in which each task has its own decoder and another where all tasks use one decoder. Based on these baselines, we establish the need to find a better decoder sharing scheme in Section \ref{res:baselines}. We then motivate the need for the PDF algorithm (Section \ref{res:prog_motive}) and show how it aids in generalization and robustness (Section \ref{res:prog_gen}). Section \ref{subsec: consist} demonstrates how the baselines and different ways of grouping affect the consistency between task predictions. Finally in Section \ref{subsec:ae_role}, we analyze the role of autoencoder.

\subsection{Baselines}
\label{res:baselines}

Since we take the setting where the encoder is always shared between all tasks, the baselines are only based on the decoder. We consider two baselines, the network with all tasks having their own decoder (referred as \textbf{Sep-De}) and the network with one decoder (referred as \textbf{One-De}) for all tasks. Since Sep-De has no shared decoder parameters, there is no task interference in the decoder. However, there is no positive transfer between tasks in the decoder and the overall inference efficiency is reduced. On the other hand, One-De shares all decoder parameters among tasks and therefore could have high task interference. However, One-De has the best inference efficiency and could also leverage positive transfer. An ideal decoder sharing scheme would surpass the performance of Sep-De by curbing task interference and leveraging positive transfer while rivaling One-De in inference efficiency.

\begin{table}[tbp]
\caption{Generalization results on in-distribution and OOD (four different natural corruption categories: noise, blur, weather, and digital). \# (M) denotes parameter count in millions. In majority of the cases, Sep-De achieves better generalization than One-De.}
\centering
\resizebox{\columnwidth}{!}{
\begin{tabular}{|c|c|ccccc|ccccc|cc|}
\toprule
\multicolumn{2}{|c|}{\multirow{2}{*}{Network}} & \multicolumn{ 5}{c|}{$\mathcal{S}$ (mIoU \%)$\uparrow$} & \multicolumn{ 5}{c|}{$\mathcal{D}$ (RMSE)$\downarrow$} & \multirow{2}{*}{FPS} & \multirow{2}{*}{\# (M)} \\ \cmidrule{3-12}
\multicolumn{ 2}{|c|}{} & IID & noise & blur & weather & digital & IID & noise & blur & weather & digital & & \\ \midrule

\multirow{2}{*}{\rotatebox[origin=c]{90}{CS}} & One-De & 72.11 & 20.28 & 48.45 & 35.86 & 60.29 & 5.36 & \textbf{15.48} & 13.20 & \textbf{11.82} & 7.60 & \textbf{32.08} & \textbf{51.56} \\ 
& Sep-De & \textbf{72.72} & \textbf{20.69} & \textbf{49.47} & \textbf{35.95} & \textbf{61.39} & \textbf{5.33} & 16.39 & \textbf{12.95} & 11.93 & \textbf{7.20} & 22.31 & 57.03 \\ \midrule

\multirow{2}{*}{\rotatebox[origin=c]{90}{NYU}} & One-De & 41.79 & \textbf{5.43} & 23.34 & 19.04 & 26.12 & 47.98 & \textbf{112.45} & \textbf{83.01} & 82.11 & 67.51 & \textbf{50.70} & \textbf{51.57} \\ 
& Sep-De & \textbf{41.99} & 4.86 & \textbf{23.53} & 19.04 & \textbf{26.70} & \textbf{47.68} & 115.36 & 85.74 & \textbf{79.75} & \textbf{64.69} & 31.24 & 57.04 \\ \bottomrule
\end{tabular}}
\label{table: iid_ood}
\end{table}


\begin{table}[tbp]
\caption{Robustness to PGD attack results reported as average over low and high $\epsilon$ bounds. In majority of the cases, Sep-De achieves more robustness over One-De.}
\centering
\begin{tabular}{|c|cc|cc|cc|cc|}
\toprule
\multirow{3}{*}{Network} & \multicolumn{4}{c|}{CS} & \multicolumn{ 4}{c|}{NYU} \\ \cmidrule{2- 9}
 & \multicolumn{2}{c|}{$\mathcal{S}$ (mIoU \%)$\uparrow$} & \multicolumn{ 2}{c|}{$\mathcal{D}$ (-RMSE)$\downarrow$} & \multicolumn{2}{c|}{$\mathcal{S}$ (mIoU \%)$\uparrow$} & \multicolumn{2}{c|}{$\mathcal{D}$ (-RMSE)$\downarrow$} \\ \cmidrule{2- 9}
 & low $\epsilon$ & high $\epsilon$ & low $\epsilon$ & high $\epsilon$ & low $\epsilon$ & high $\epsilon$ & low $\epsilon$ & high $\epsilon$ \\ \midrule
One-De & 45.04 & 6.91 & \textbf{15.16} & 46.32 & 19.34 & 2.73 & 110.86 & 309.56 \\ 
Sep-De & \textbf{45.95} & \textbf{7.92} & 15.24 & \textbf{45.14} & \textbf{20.36} & \textbf{3.29} & \textbf{108.07} & \textbf{303.61} \\ \bottomrule
\end{tabular}
\label{table: robustness}
\end{table}

Table \ref{table: iid_ood} presents the results obtained with both the baselines on the Cityscapes and the NYUv2 datasets. We report only the segmentation (S) and depth (D) results for brevity and because these tasks are the main tasks of concern. Except for a few cases, we see that Sep-De always results in a performance gain relative to One-De. We attribute this gain to the absence of task interference between all tasks in the decoder parameter space. However, this gain in performance of Sep-De comes at the cost of reduced inference efficiency as is evident from the slower FPS and higher parameter count in comparison to One-De. In Table \ref{table: robustness}, we see that Sep-De also achieves better robustness to PGD attack in majority of the cases. The PDF algorithm intends to retain or improve over the generalization and robustness gain of Sep-De while finding a balance in inference efficiency.

\subsection{Why Progressive Decoder Fusion?}
\label{res:prog_motive}

In Section \ref{res:baselines}, we established the baselines and saw that Sep-De results in a performance gain. To build upon this performance gain by leveraging positive transfer and increasing inference efficiency, we intend to curb task interference by only sharing similar decoder representations. Based on the 5$\times$5 CKA similarity matrix, we group the tasks at candidate stage 1. The grouping algorithm (detailed in Algorithm \ref{algo:1}) first lists all possible combinations of task groupings. Out of all these combinations, the one which has the highest value defined by the average similarity across all tasks in the grouping is picked. For instance in the Cityscapes dataset, the best grouping obtained using Sep-De is [$\mathcal{S}=0.86$, $\mathcal{E}=0.86$], [$\mathcal{D}=0.76$, $\mathcal{N}=0.76$], [$\mathcal{A}=0$] with overall value of 0.65. In the picked grouping, if a two-task group has a value of less than 0.5, the tasks in the group are split back into their own branches. After grouping tasks in candidate stage 1, we retrain this new network until convergence. In the next stage, grouping is only done based on the $t\times t$ for the $t$ tasks sharing the same group in the previous stage. We continue this process until grouping is done in all stages. 


\begin{figure}
  \begin{minipage}[c]{0.6\textwidth}
    \includegraphics[width=1\linewidth]{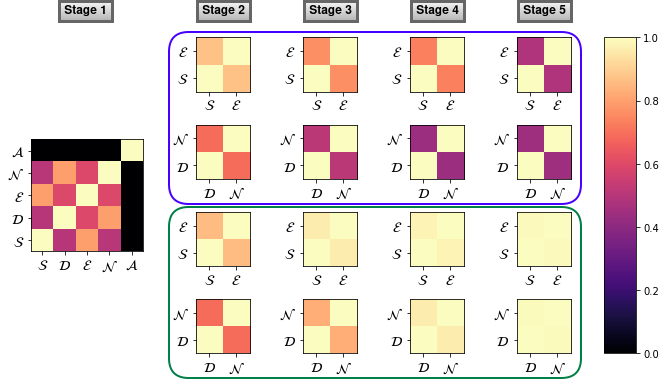}
  \end{minipage}\hfill
  \begin{minipage}[c]{0.35\textwidth}
    \caption{CKA similarity matrix at different decoder stages obtained from offline (blue box) and progressive (green box) grouping on Cityscapes dataset. As we move towards the right, decoder representations become increasingly similar with progressive grouping but increasingly dissimilar with offline grouping.} \label{fig:cka_heatmap}
  \end{minipage}
\end{figure}

The changes in the similarity between tasks are depicted by visualizing the CKA similarity matrix in Figure \ref{fig:cka_heatmap} for the Cityscapes dataset. The leftmost plot shows the 5$\times$5 pairwise similarity across the 5 tasks in Sep-De. In offline grouping, tasks are fused at all stages based on these inter-task similarities. This grouping is similar to how \cite{BranchedVand} group tasks in the encoder. In offline grouping (blue box), at stage 2, the two task groups picked based on stage 1 similarities are [$\mathcal{S}$, $\mathcal{E}$] and [$\mathcal{D}$, $\mathcal{N}$]. The similarity between the same task groups is visualized across the rest of the stages. These results show that in offline grouping, tasks become increasingly dissimilar across stages (from left to right).

On the other hand, in progressive grouping (green box), the tasks become increasingly similar across stages. This observation shows that once tasks are grouped at a stage, the similarity between grouped tasks in the subsequent stages increases. This effect supports the hypothesis that task similarities change based on where they branch in the network. However, offline grouping does not consider this effect leading to groupings at stages 2 to 5 which might not be ideal. This possibility serves as a motivation to set up and evaluate the PDF algorithm. We train a network with the decoder sharing scheme obtained with offline grouping, referred to as \textbf{Offline} and compare with \textbf{PDF} which is the final model obtained with the PDF algorithm in Section \ref{res:prog_gen}.

\subsection{Generalization and Robustness}
\label{res:prog_gen}
Results in Section \ref{res:prog_motive} suggest that grouping task decoders progressively at each stage could result in a decoder sharing scheme that generalizes better and is also more robust. This expectation stems from the nature of the PDF algorithm which only groups similar tasks at every candidate stage. Since only similar tasks are grouped, positive transfer could improve while reducing task interference leading to richer representations that leverage the strengths of each task.

 

\begin{table}[tbp]
\caption{Generalization and robustness results of Sep-De, Offline and PDF networks. Results of IID and OOD (noise, blue, weather, and digital) generalization and robustness (low $\epsilon$ and high $\epsilon$) are reported as percentage improvements over One-De. \# (M) denotes parameter count in millions. PDF network outperforms Offline network in most of the cases while being more inference efficient.}
\centering
\begin{tabular}{|c|c|c|cccc|cc|cc|}
\toprule
\multicolumn{ 2}{|c|}{Network} & IID & noise & blur & weather & digital & low $\epsilon$ & high $\epsilon$ & FPS & \# (M) \\ \midrule
\multirow{3}{*}{\rotatebox[origin=c]{90}{CS}} & Sep-De & 0.66 & -1.90 & 2.00 & -0.35 & 3.57 & 0.75 & 8.62 & 22.31 & 57.03 \\ \cmidrule{ 2- 11}
& Offline & 0.71 & -3.54 & 0.28 & -0.26 & \textbf{2.05} & \textbf{0.73} & \textbf{10.35} & 24.19 & 54.57 \\ 
 & PDF & \textbf{1.00} & \textbf{-0.10} & \textbf{1.36} & \textbf{-0.01} & 1.13 & 0.70 & 9.97 & \textbf{25.44} & \textbf{54.30} \\ \midrule
 
\multirow{3}{*}{\rotatebox[origin=c]{90}{NYU}} & Sep-De & 0.55 & -6.51 & -1.24 & 1.43 & 3.19 & 3.89 & 11.13 & 31.24 & 57.04 \\ \cmidrule{ 2- 11}
& Offline & 0.80 & -5.95 & \textbf{1.93} & 0.48 & 1.61 & 2.96 & 15.05 & 33.02 & 55.86 \\ 
 & PDF & \textbf{1.38} & \textbf{-4.04} & 1.46 & \textbf{1.88} & \textbf{3.62} & \textbf{4.11} & \textbf{16.16} & \textbf{34.74} & \textbf{55.67} \\ \bottomrule
\end{tabular}
\label{table: pdf5-offline}
\end{table}

\begin{figure}[tbhp]
  \centering
  \includegraphics[width=.7\linewidth]{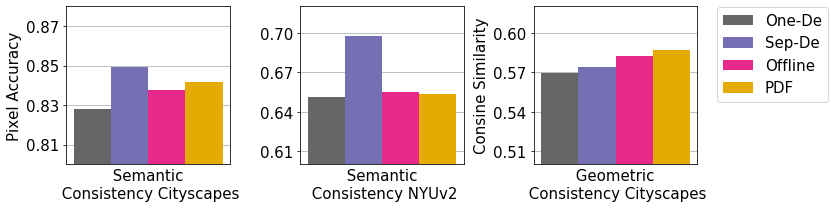}
   \caption{Semantic and geometric consistency results. PDF network provides more consistent predictions than the offline network except in NYUv2 semantic consistency. Compared to Sep-De, PDF results in higher consistency only in cityscapes geometric consistency.}
\label{fig:consistency}
\end{figure}

The groupings done at different decoder stages using progressive and offline grouping for both datasets are provided in Table \ref{table:all_groups} in Appendix. For Cityscapes in stage 1, the task groups are [$\mathcal{S}$, $\mathcal{E}$], [$\mathcal{D}$, $\mathcal{N}$], [$\mathcal{A}$]. This grouping indicates that the tasks $\mathcal{S}$, $\mathcal{E}$ are to be combined in stage 1. Likewise, $\mathcal{D}$, $\mathcal{N}$ are combined while $\mathcal{A}$ has its own decoder. We see that the grouping at stages 4 and 5 differ between PDF algorithm and offline grouping. Therefore, these two ways of grouping result in different final models. In NYUv2 the groupings differ right from stage 1. 

Table \ref{table: pdf5-offline} details the generalization and robustness results of PDF and Offline networks. The numbers are reported as the average relative improvement of the segmentation and depth tasks over One-De using Equation \ref{eq:rel_imp}. \footnote{Relative depth RMSE is subtracted as lower depth RMSE is better.}
\begin{equation}
    \frac{1}{2} * \left(\frac{(\mathcal{S}_{{net}} - \mathcal{S}_{\textnormal{One-De}})}{\mathcal{S}_{\textnormal{One-De}}} - \frac{(\mathcal{D}_{{net}} - \mathcal{D}_{\textnormal{One-De}})}{\mathcal{D}_{\textnormal{One-De}}} \right)* 100, ~~~~\textnormal{where} ~ {net} \in \{\textnormal{Sep-De}, \textnormal{PDF}, \textnormal{Offline}\}
    \label{eq:rel_imp}
\end{equation}

The PDF network surpasses the Offline network in a majority of cases. These results support the PDF algorithm and show that changing grouping decisions in a stage based on a fully trained network with tasks grouped up to the previous stage is helpful. We also see that PDF is more inference efficient than the Offline network as is evident from the FPS and number of parameters. Overall, the decoder sharing scheme of PDF mostly achieves better generalization and robustness than the Offline network while being more inference efficient.

\subsection{Inter-task prediction consistency}
\label{res:consistency}
The consistency between predictions of different tasks could be seen as an indicator of task interference. When the predictions are consistent, tasks could have shared more complementary information rather than conflicting information. Thus, higher prediction consistency may indicate reduced task interference. We explore this possibility and look at the consistency between two sets of predictions. 

The consistencies of different models are visualized in Figure \ref{fig:consistency}\footnote{We do not include NYUv2 geometric consistency as we have used the precomputed surface normals for training and have not reproduced computation of normals from depth using the NYUv2 toolbox.}. In Cityscapes, in both semantic and geometric consistency, PDF provides more consistent predictions compared to the Offline network suggesting that it has leveraged positive transfer. However, in NYUv2 semantic consistency, task interference likely hasn't been reduced enough as PDF falls slightly behind the Offline network. PDF falls behind Sep-De in semantic consistency in both datasets suggesting that there is room to further curb task interference. These consistencies likely only provide weak evidence for task interference and further evaluations would be required to draw insights from consistency.

\subsection{Role of Autoencoder}
\label{subsec:ae_role}
The autoencoder learns to reconstruct the input image with the aid of an MSE loss and would require learning a different set of features that takes into account the varied aspects of the input scene. One evidence for this requirement is seen in the CKA similarity matrix obtained with Sep-De in Figure \ref{fig:cka_heatmap} (leftmost heatmap). The autoencoder feature representations at decoder stage 1 are different from all other task representations in both datasets. Likewise, a different set of features could also be learnt in the encoder. These different features enforced by the autoencoder are likely to be more general to facilitate image reconstruction. Since all tasks branch from a potentially more general encoder representation, the generalization and robustness of tasks could have improved.

\begin{table}[tbp]
\caption{The IID and OOD (noise, blue, weather, and digital) generalization results of PDF network with and without autoencoder ($\mathcal{A}$). Overall, autoencoder results in generalization improvements except in Cityscapes depth.}
\centering
\resizebox{\columnwidth}{!}{
\begin{tabular}{|c|c|ccccc|ccccc|}
\toprule
\multicolumn{2}{|c|}{\multirow{2}{*}{Network}} & \multicolumn{5}{c|}{$\mathcal{S}$ (mIoU \%)$\uparrow$} & \multicolumn{5}{c|}{$\mathcal{D}$ (RMSE) $\downarrow$}  \\ \cmidrule{ 3- 12}
\multicolumn{2}{|c|}{} & IID & noise & blur & weather & digital & IID & noise & blur & weather & digital \\ \midrule
\multirow{2}{*}{\rotatebox[origin=c]{90}{CS}} & PDF & \textbf{72.92} & \textbf{19.72} & \textbf{49.19} & 36.68 & \textbf{60.98} & \textbf{5.31} & 15.09 & 13.04 & 12.09 & 7.52 \\ 
& PDF-No-$\mathcal{A}$ & 72.17 & 18.56 & 48.76 & \textbf{36.72} & 60.88 & 5.32 & \textbf{13.84} & \textbf{12.99} & \textbf{11.91} & \textbf{7.29} \\  \midrule
\multirow{2}{*}{\rotatebox[origin=c]{90}{NYU}} & PDF & \textbf{42.28} & 5.04 & \textbf{24.27} & 18.83 & \textbf{27.14} & \textbf{47.22} & \textbf{113.53} & \textbf{83.89} & \textbf{78.14} & \textbf{65.23} \\ 
& PDF-No-$\mathcal{A}$ & 42.09 & \textbf{5.49} & 23.98 & \textbf{19.01} & 26.79 & 47.39 & 134.45 & 86.86 & 80.31 & 65.31 \\ \bottomrule
\end{tabular}}
\label{table: ae_gen}
\end{table}

\begin{table}[tbp]
\caption{Robustness to PGD attack results reported as average over low and high $\epsilon$ bounds for PDF network with and without autoencoder ($\mathcal{A}$). Overall, autoencoder results in robustness improvements.}
\centering
\begin{tabular}{|c|cc|cc|cc|cc|}
\toprule
\multirow{3}{*}{Network} & \multicolumn{ 4}{c|}{CS} & \multicolumn{ 4}{c|}{NYU} \\ \cmidrule{2- 9}
 & \multicolumn{2}{c|}{$\mathcal{S}$ (mIoU \%) $\uparrow$} & \multicolumn{ 2}{c|}{$\mathcal{D}$ (RMSE) $\downarrow$} & \multicolumn{ 2}{c|}{$\mathcal{S}$ (mIoU \%) $\uparrow$} & \multicolumn{2}{c|}{$\mathcal{D}$ (RMSE) $\downarrow$} \\ \cmidrule{2-9}
 & low $\epsilon$ & high $\epsilon$ & low $\epsilon$ & high $\epsilon$ & low $\epsilon$ & high $\epsilon$ & low $\epsilon$ & high $\epsilon$ \\ \midrule
PDF & \textbf{45.75} & 8.07 & \textbf{15.19} & 44.93 & \textbf{20.38} & \textbf{3.58} & \textbf{107.71} & \textbf{304.94} \\ 
PDF-No-$\mathcal{A}$ & 45.65 & \textbf{8.29} & 15.29 & \textbf{44.82} & 20.34 & 3.56 & 121.67 & 323.66 \\ \bottomrule
\end{tabular}
\label{table: ae_robust}
\end{table}

We evaluate the effect of the autoencoder by comparing the generalization and robustness of the PDF network with and without the autoencoder. Table \ref{table: ae_gen} shows that adding an autoencoder generally improves or retains the generalization except in Cityscapes depth estimation. Table \ref{table: ae_robust} shows that autoencoder leads to robustness improvements in most cases with considerable improvements in NYUv2 depth estimation. Overall, we note that autoencoder provides reasonable generalization and robustness improvements in most cases across tasks and datasets.


\section{Limitations}

The datasets used have limited training images likely leading to a small difference between Sep-De and One-De in our experimental setup. This small performance gap stands as a hindrance to sufficiently gauge the utility of the PDF algorithm. With more complex and larger datasets, this issue could be overcome. Additionally, the training time required to arrive at the final network is higher than normal as multiple retraining from initialization is required rendering the method infeasible in situations where only limited training resources is available. Another limitation is that we assume that the encoder is always shared among all tasks. We intend to explore a more training time efficient approach while also considering the encoder for fusion in our future work.

\section{Conclusion}
To curb task interference in the decoder while increasing positive transfer and improving inference efficiency, we proposed progressive decoder fusion starting from each task having its own decoder and grouping tasks in subsequent decoder stages in every progression step. At every step in this progression, tasks are grouped based on CKA similarity and the obtained network is retrained. We showed that the proposed PDF algorithm leads to a network that surpasses the offline grouped network in inference efficiency, generalization, and robustness. With this work, we wish to have reinforced interest in the research community to curb task interference and expose the full potential of MTNs.

\bibliography{egbib}
\bibliographystyle{collas2022_conference}

\appendix
\section{Appendix}

\subsection{Additional related works}

In addition to the related works presented in the main paper, we differentiate our work with other research directions for completeness. An alternate line of work concerns fusing the learned task specific features using additional parameters to enhance task interaction \citep{8578175, Vandenhende2020MTINetMT}. Fusion has also been done at the input, where data from multiple modalities are fused together \citep{8954034}. Unlike these works, we fuse the parameters of the networks themselves. \cite{7423818} consider a training scheme in which they alternate between multi-task learning and finding task-relatedness. In a training iteration, only the parameters of related tasks are trained thereby reducing interference from the learning objectives of other tasks. Similarly, notions of task relatedness have been proposed in the context of multitask learning \citep{Standley2020WhichTS, Lu2021TaskologyUT} and transfer learning \citep{Zamir_2018_CVPR, Dwivedi2019RepresentationSA} to help improve generalization. Unlike these works, we group tasks and directly fuse task-specific parameters.

\subsection{Centered Kernel Alingment (CKA)}
\label{section:CKA_detailed}

CKA provides a way to compare different representations learned by a neural network. We use CKA to compare the output features from different tasks at a decoder stage. CKA is computed between every pair of task representations X and Y obtained using N images. We first take the mean across spatial locations\footnote{Note that alternatively, mean across channels can be taken followed by flattening along the spatial dimensions.}. We use the unbiased CKA estimation method provided by \cite{nguyen2021do} to first calculate $HSIC_1$ using Equation \ref{eq:hsic_eq} where K and L represent Gram matrices $XX^T$ and $YY^T$, respectively. $HSIC_1$ is used to calculate CKA with the aid of Equation \ref{eq:cka_eq} where $\tilde{K}$ and $\tilde{L}$ are modified versions of K and L with the diagonal entries made zero.

\begin{equation}
    HSIC_1(K, L) = \frac{1}{n(n-3)}\left(tr(\tilde{K}\tilde{L}) + \frac{1^T\tilde{K}11^T\tilde{L}1}{(n - 1)(n - 2)} - \frac{2}{n - 2}1^T\tilde{K}\tilde{L}1\right)
    \label{eq:hsic_eq}
\end{equation}

\begin{equation}
    CKA = \frac{HSIC_1(K,L)}{\sqrt{HSIC_1(K,L)} \sqrt{HSIC_1(K,L)}}
    \label{eq:cka_eq}
\end{equation}

\cite{BranchedVand} use Representation Similarity Analysis (RSA). Similar to CKA, the mean across either the channel dimensions or the spatial dimensions is taken to obtain a vector for each input image. Next, a Representation Dissimilarity Matrix (RDM) of dimension N$\times$N for each representation X and Y is calculated as 1 - Person correlation between all pairs of N feature vectors. From both the RDM matrices, a flattened vector from the upper triangle matrix is obtained. The Spearman's correlation between the two flattened vector represents the RSA\footnote{Note there could be other versions of RSA with different formulations.} between representations X and Y.

\subsection{Additional training details}
We use UniNet architecture for the experiments and we refer to D6-D2 as stages 1-5. We use ResNet50 as backbone architecture to provide the encoder features E2 to E5 while E6 and E7 are obtained using the first two ResNet18 encoder stages. The encoder extracts the features from the input image and provides a feature representation with low spatial resolution. The decoder maps the extracted encoder features to a higher spatial resolution which is in turn mapped to the corresponding output space by each task head. To train the edge detection task in both datasets, we transform the semantic segmentation ground truth to edge detection ground truth. For the surface normals in Cityscapes, we use the depth ground truth transformed to surface normals ground truth for training. In NYUv2, we use the precomputed surface normals ground truth available in the dataset. Class balanced cross entropy loss, RMSE loss, class balanced binary cross entropy loss, L1 loss and MSE loss is used to train semantic segmentation, depth estimation, edge detection, surface normals and autoencoder, respectively.

\subsection{Obtained task groupings}
The obtained task groupings for both datasets at different stages using the PDF algorithm and Offline grouping are listed in Table \ref{table:all_groups}. Additionally, the effect on the resultant task groupings while changing the grouping threshold is presented in Table \ref{table:all_groups_all}.

\begin{table}[h]
\caption{Task groupings for both datasets at all candidate stages obtained with PDF algorithm and offline grouping. Blue highlighted cells show the stages that offline grouping deviates from progressive grouping.}
\label{table:all_groups}
\centering
\begin{tabular}{|c|c|c|c|c|}
\toprule
 Stage & \multicolumn{ 2}{c|}{Cityscapes} & \multicolumn{ 2}{c|}{NYUv2} \\ \cmidrule{2-5}
 & PDF & Offline & PDF & Offline \\ \midrule
1,2,3 & [$\mathcal{S}$, $\mathcal{E}$], [$\mathcal{D}$, $\mathcal{N}$], [$\mathcal{A}$] & [$\mathcal{S}$, $\mathcal{E}$], [$\mathcal{D}$, $\mathcal{N}$], [$\mathcal{A}$] & [$\mathcal{S}$, $\mathcal{E}$], [$\mathcal{D}$], [$\mathcal{N}$], [$\mathcal{A}$] & [$\mathcal{S}$, $\mathcal{E}$], [$\mathcal{D}$], [$\mathcal{N}$], [$\mathcal{A}$] \\ 
4 & [$\mathcal{S}$, $\mathcal{E}$], [$\mathcal{D}$, $\mathcal{N}$], [$\mathcal{A}$] & \cellcolor{blue!25}[$\mathcal{S}$, $\mathcal{E}$], [$\mathcal{D}$], [$\mathcal{N}$], [$\mathcal{A}$] & [$\mathcal{S}$, $\mathcal{E}$], [$\mathcal{D}$], [$\mathcal{N}$], [$\mathcal{A}$] & \cellcolor{blue!25}[$\mathcal{S}$], [$\mathcal{E}$], [$\mathcal{D}$], [$\mathcal{N}$], [$\mathcal{A}$] \\ 
5 & [$\mathcal{S}$, $\mathcal{E}$], [$\mathcal{D}$, $\mathcal{N}$], [$\mathcal{A}$] & \cellcolor{blue!25}[$\mathcal{S}$], [$\mathcal{E}$], [$\mathcal{D}$], [$\mathcal{N}$], [$\mathcal{A}$] & [$\mathcal{S}$, $\mathcal{E}$], [$\mathcal{D}$], [$\mathcal{N}$], [$\mathcal{A}$] & \cellcolor{blue!25}[$\mathcal{S}$], [$\mathcal{E}$], [$\mathcal{D}$], [$\mathcal{N}$], [$\mathcal{A}$] \\ 
\bottomrule
\end{tabular}
\end{table}

\begin{table}[h]
\caption{Task groupings for both datasets obtained with PDF algorithm and offline grouping using different thresholds for grouping (denoted as $\tau$). Blue highlighted cells show the stages where the offline grouping deviates from progressive grouping.}
\label{table:all_groups_all}
\centering
\resizebox{\columnwidth}{!}{
\begin{tabular}{|c|c|c|c|c|c|}
\toprule
Stage & \multirow{2}{*}{$\tau$} & \multicolumn{ 2}{c|}{Cityscapes} & \multicolumn{ 2}{c|}{NYUv2} \\ \cmidrule{3-6}
& & PDF & Offline & PDF & Offline \\ \midrule
1,2,3,4,5 & 0.0 & [$\mathcal{S}$, $\mathcal{E}$], [$\mathcal{D}$, $\mathcal{N}$], [$\mathcal{A}$] & [$\mathcal{S}$, $\mathcal{E}$], [$\mathcal{D}$, $\mathcal{N}$], [$\mathcal{A}$] & [$\mathcal{S}$, $\mathcal{E}$], [$\mathcal{D}$, $\mathcal{N}$], [$\mathcal{A}$] & [$\mathcal{S}$, $\mathcal{E}$], [$\mathcal{D}$, $\mathcal{N}$], [$\mathcal{A}$] \\ 
1,2,3,4,5 & 0.1 & [$\mathcal{S}$, $\mathcal{E}$], [$\mathcal{D}$, $\mathcal{N}$], [$\mathcal{A}$] & [$\mathcal{S}$, $\mathcal{E}$], [$\mathcal{D}$, $\mathcal{N}$], [$\mathcal{A}$] & [$\mathcal{S}$, $\mathcal{E}$], [$\mathcal{D}$, $\mathcal{N}$], [$\mathcal{A}$] & [$\mathcal{S}$, $\mathcal{E}$], [$\mathcal{D}$, $\mathcal{N}$], [$\mathcal{A}$] \\ 
1,2,3,4,5 & 0.2 & [$\mathcal{S}$, $\mathcal{E}$], [$\mathcal{D}$, $\mathcal{N}$], [$\mathcal{A}$] & [$\mathcal{S}$, $\mathcal{E}$], [$\mathcal{D}$, $\mathcal{N}$], [$\mathcal{A}$] & [$\mathcal{S}$, $\mathcal{E}$], [$\mathcal{D}$, $\mathcal{N}$], [$\mathcal{A}$] & [$\mathcal{S}$, $\mathcal{E}$], [$\mathcal{D}$, $\mathcal{N}$], [$\mathcal{A}$] \\ \midrule

1,2,3,4 & \multirow{2}{*}{0.3} & [$\mathcal{S}$, $\mathcal{E}$], [$\mathcal{D}$, $\mathcal{N}$], [$\mathcal{A}$] & [$\mathcal{S}$, $\mathcal{E}$], [$\mathcal{D}$, $\mathcal{N}$], [$\mathcal{A}$] & [$\mathcal{S}$, $\mathcal{E}$], [$\mathcal{D}$, $\mathcal{N}$], [$\mathcal{A}$] & [$\mathcal{S}$, $\mathcal{E}$], [$\mathcal{D}$, $\mathcal{N}$], [$\mathcal{A}$] \\ 
5 &  & [$\mathcal{S}$, $\mathcal{E}$], [$\mathcal{D}$, $\mathcal{N}$], [$\mathcal{A}$] & [$\mathcal{S}$, $\mathcal{E}$], [$\mathcal{D}$, $\mathcal{N}$], [$\mathcal{A}$] & \cellcolor{blue!25}[$\mathcal{S}$, $\mathcal{E}$], [$\mathcal{D}$, $\mathcal{N}$], [$\mathcal{A}$] & \cellcolor{blue!25}[$\mathcal{S}$, $\mathcal{E}$], [$\mathcal{D}$], [$\mathcal{N}$], [$\mathcal{A}$] \\ \midrule

1,2,3,4 & \multirow{2}{*}{0.4} & [$\mathcal{S}$, $\mathcal{E}$], [$\mathcal{D}$, $\mathcal{N}$], [$\mathcal{A}$] & [$\mathcal{S}$, $\mathcal{E}$], [$\mathcal{D}$, $\mathcal{N}$], [$\mathcal{A}$] & [$\mathcal{S}$, $\mathcal{E}$], [$\mathcal{D}$], [$\mathcal{N}$], [$\mathcal{A}$] & [$\mathcal{S}$, $\mathcal{E}$], [$\mathcal{D}$], [$\mathcal{N}$], [$\mathcal{A}$] \\
5 &  & [$\mathcal{S}$, $\mathcal{E}$], [$\mathcal{D}$, $\mathcal{N}$], [$\mathcal{A}$] & [$\mathcal{S}$, $\mathcal{E}$], [$\mathcal{D}$, $\mathcal{N}$], [$\mathcal{A}$] & \cellcolor{blue!25}[$\mathcal{S}$, $\mathcal{E}$], [$\mathcal{D}$], [$\mathcal{N}$], [$\mathcal{A}$] & \cellcolor{blue!25}[$\mathcal{S}$], [$\mathcal{E}$], [$\mathcal{D}$], [$\mathcal{N}$], [$\mathcal{A}$] \\ \midrule

1,2,3 & \multirow{2}{*}{0.5} & [$\mathcal{S}$, $\mathcal{E}$], [$\mathcal{D}$, $\mathcal{N}$], [$\mathcal{A}$] & [$\mathcal{S}$, $\mathcal{E}$], [$\mathcal{D}$, $\mathcal{N}$], [$\mathcal{A}$] & [$\mathcal{S}$, $\mathcal{E}$], [$\mathcal{D}$], [$\mathcal{N}$], [$\mathcal{A}$] & [$\mathcal{S}$, $\mathcal{E}$], [$\mathcal{D}$], [$\mathcal{N}$], [$\mathcal{A}$] \\ 
4 &  & \cellcolor{blue!25}[$\mathcal{S}$, $\mathcal{E}$], [$\mathcal{D}$, $\mathcal{N}$], [$\mathcal{A}$] & \cellcolor{blue!25}[$\mathcal{S}$, $\mathcal{E}$], [$\mathcal{D}$], [$\mathcal{N}$], [$\mathcal{A}$] & \cellcolor{blue!25}[$\mathcal{S}$, $\mathcal{E}$], [$\mathcal{D}$], [$\mathcal{N}$], [$\mathcal{A}$] & \cellcolor{blue!25}[$\mathcal{S}$], [$\mathcal{E}$], [$\mathcal{D}$], [$\mathcal{N}$], [$\mathcal{A}$] \\ 
5 &  & \cellcolor{blue!25}[$\mathcal{S}$, $\mathcal{E}$], [$\mathcal{D}$, $\mathcal{N}$], [$\mathcal{A}$] & \cellcolor{blue!25}[$\mathcal{S}$], [$\mathcal{E}$], [$\mathcal{D}$], [$\mathcal{N}$], [$\mathcal{A}$] & \cellcolor{blue!25}[$\mathcal{S}$, $\mathcal{E}$], [$\mathcal{D}$], [$\mathcal{N}$], [$\mathcal{A}$] & \cellcolor{blue!25}[$\mathcal{S}$], [$\mathcal{E}$], [$\mathcal{D}$], [$\mathcal{N}$], [$\mathcal{A}$] \\ \midrule

1,2 & \multirow{2}{*}{0.6} & [$\mathcal{S}$, $\mathcal{E}$], [$\mathcal{D}$, $\mathcal{N}$], [$\mathcal{A}$] & [$\mathcal{S}$, $\mathcal{E}$], [$\mathcal{D}$, $\mathcal{N}$], [$\mathcal{A}$] & [$\mathcal{S}$, $\mathcal{E}$], [$\mathcal{D}$], [$\mathcal{N}$], [$\mathcal{A}$] & [$\mathcal{S}$, $\mathcal{E}$], [$\mathcal{D}$], [$\mathcal{N}$], [$\mathcal{A}$] \\ 
3,4 &  & \cellcolor{blue!25}[$\mathcal{S}$, $\mathcal{E}$], [$\mathcal{D}$, $\mathcal{N}$], [$\mathcal{A}$] & \cellcolor{blue!25}[$\mathcal{S}$, $\mathcal{E}$], [$\mathcal{D}$], [$\mathcal{N}$], [$\mathcal{A}$] & \cellcolor{blue!25}[$\mathcal{S}$, $\mathcal{E}$], [$\mathcal{D}$], [$\mathcal{N}$], [$\mathcal{A}$] & \cellcolor{blue!25}[$\mathcal{S}$], [$\mathcal{E}$], [$\mathcal{D}$], [$\mathcal{N}$], [$\mathcal{A}$] \\ 
5 &  & \cellcolor{blue!25}[$\mathcal{S}$, $\mathcal{E}$], [$\mathcal{D}$, $\mathcal{N}$], [$\mathcal{A}$] & \cellcolor{blue!25}[$\mathcal{S}$], [$\mathcal{E}$], [$\mathcal{D}$], [$\mathcal{N}$], [$\mathcal{A}$] & \cellcolor{blue!25}[$\mathcal{S}$, $\mathcal{E}$], [$\mathcal{D}$], [$\mathcal{N}$], [$\mathcal{A}$] & \cellcolor{blue!25}[$\mathcal{S}$], [$\mathcal{E}$], [$\mathcal{D}$], [$\mathcal{N}$], [$\mathcal{A}$] \\ \midrule

1 & \multirow{2}{*}{0.7} & [$\mathcal{S}$, $\mathcal{E}$], [$\mathcal{D}$, $\mathcal{N}$], [$\mathcal{A}$] & [$\mathcal{S}$, $\mathcal{E}$], [$\mathcal{D}$, $\mathcal{N}$], [$\mathcal{A}$] & [$\mathcal{S}$], [$\mathcal{E}$], [$\mathcal{D}$], [$\mathcal{N}$], [$\mathcal{A}$] & [$\mathcal{S}$], [$\mathcal{E}$], [$\mathcal{D}$], [$\mathcal{N}$], [$\mathcal{A}$] \\ 
2,3,4 &  & [$\mathcal{S}$, $\mathcal{E}$], [$\mathcal{D}$], [$\mathcal{N}$], [$\mathcal{A}$] & [$\mathcal{S}$, $\mathcal{E}$], [$\mathcal{D}$], [$\mathcal{N}$], [$\mathcal{A}$] & [$\mathcal{S}$], [$\mathcal{E}$], [$\mathcal{D}$], [$\mathcal{N}$], [$\mathcal{A}$] & [$\mathcal{S}$], [$\mathcal{E}$], [$\mathcal{D}$], [$\mathcal{N}$], [$\mathcal{A}$] \\ 
5 &  & \cellcolor{blue!25}[$\mathcal{S}$, $\mathcal{E}$], [$\mathcal{D}$], [$\mathcal{N}$], [$\mathcal{A}$] & \cellcolor{blue!25}[$\mathcal{S}$], [$\mathcal{E}$], [$\mathcal{D}$], [$\mathcal{N}$], [$\mathcal{A}$] & [$\mathcal{S}$], [$\mathcal{E}$], [$\mathcal{D}$], [$\mathcal{N}$], [$\mathcal{A}$] & [$\mathcal{S}$], [$\mathcal{E}$], [$\mathcal{D}$], [$\mathcal{N}$], [$\mathcal{A}$] \\ \midrule

1 & \multirow{2}{*}{0.8} & [$\mathcal{S}$, $\mathcal{E}$], [$\mathcal{D}$, $\mathcal{N}$], [$\mathcal{A}$] & [$\mathcal{S}$, $\mathcal{E}$], [$\mathcal{D}$, $\mathcal{N}$], [$\mathcal{A}$] & [$\mathcal{S}$], [$\mathcal{E}$], [$\mathcal{D}$], [$\mathcal{N}$], [$\mathcal{A}$] & [$\mathcal{S}$], [$\mathcal{E}$], [$\mathcal{D}$], [$\mathcal{N}$], [$\mathcal{A}$] \\ 
2 &  & [$\mathcal{S}$, $\mathcal{E}$], [$\mathcal{D}$], [$\mathcal{N}$], [$\mathcal{A}$] & [$\mathcal{S}$, $\mathcal{E}$], [$\mathcal{D}$], [$\mathcal{N}$], [$\mathcal{A}$] & [$\mathcal{S}$], [$\mathcal{E}$], [$\mathcal{D}$], [$\mathcal{N}$], [$\mathcal{A}$] & [$\mathcal{S}$], [$\mathcal{E}$], [$\mathcal{D}$], [$\mathcal{N}$], [$\mathcal{A}$] \\ 
3,4,5 &  & \cellcolor{blue!25}[$\mathcal{S}$, $\mathcal{E}$], [$\mathcal{D}$], [$\mathcal{N}$], [$\mathcal{A}$] & \cellcolor{blue!25}[$\mathcal{S}$], [$\mathcal{E}$], [$\mathcal{D}$], [$\mathcal{N}$], [$\mathcal{A}$] & [$\mathcal{S}$], [$\mathcal{E}$], [$\mathcal{D}$], [$\mathcal{N}$], [$\mathcal{A}$] & [$\mathcal{S}$], [$\mathcal{E}$], [$\mathcal{D}$], [$\mathcal{N}$], [$\mathcal{A}$] \\ \midrule

1,2,3,4,5 & 0.9 & [$\mathcal{S}$], [$\mathcal{E}$], [$\mathcal{D}$], [$\mathcal{N}$], [$\mathcal{A}$] & [$\mathcal{S}$], [$\mathcal{E}$], [$\mathcal{D}$], [$\mathcal{N}$], [$\mathcal{A}$] & [$\mathcal{S}$], [$\mathcal{E}$], [$\mathcal{D}$], [$\mathcal{N}$], [$\mathcal{A}$] & [$\mathcal{S}$], [$\mathcal{E}$], [$\mathcal{D}$], [$\mathcal{N}$], [$\mathcal{A}$] \\ 
1,2,3,4,5 & 1.0 & [$\mathcal{S}$], [$\mathcal{E}$], [$\mathcal{D}$], [$\mathcal{N}$], [$\mathcal{A}$] & [$\mathcal{S}$], [$\mathcal{E}$], [$\mathcal{D}$], [$\mathcal{N}$], [$\mathcal{A}$] & [$\mathcal{S}$], [$\mathcal{E}$], [$\mathcal{D}$], [$\mathcal{N}$], [$\mathcal{A}$] & [$\mathcal{S}$], [$\mathcal{E}$], [$\mathcal{D}$], [$\mathcal{N}$], [$\mathcal{A}$] \\ 
\bottomrule
\end{tabular}}
\end{table}

\subsection{NYUv2 CKA similarity matrices}

\begin{figure} [h]
  \centering
  \includegraphics[width=.8\linewidth]{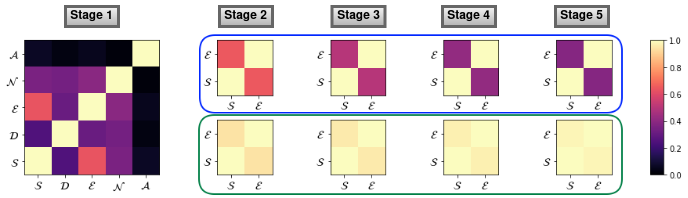}
   \caption{CKA similarity matrix at different decoder stages obtained with offline grouping (blue box) and with progressive (green box) in the NYUv2 dataset. As we move towards the right, decoder representations become increasingly similar with progressive grouping but increasingly dissimilar with offline grouping.}
\label{fig:cka_heatmap_nyuv2}
\end{figure}

The CKA similarity matrix obtained at different stages is provided in Figure \ref{fig:cka_heatmap_nyuv2}. We only visualize the similarities between $\mathcal{S}$ and $\mathcal{E}$ through different stages as all other tasks split into their own branch right from stage 1. Similar to Cityscapes similarities in Figure \ref{fig:cka_heatmap}, as we move towards the right, the decoder representations become increasingly similar with progressive grouping (green box) but increasingly dissimilar with offline grouping (blue box).

\subsection{Detailed results of all networks}
We report the mean and standard deviation of generalization results in Table \ref{table:mean_gen} and Table \ref{table:std_gen}, respectively. Likewise the mean and standard deviation of robustness results are reported in Table \ref{table:mean_robust} and Table \ref{table:std_robust}, respectively. PDF-1 to PDF-5 refer to the various networks obtained at different stages of the decoder progression. Table \ref{table:mean_and_std_gen} and Table \ref{table:mean_and_std_robust} summarize the generalization and robustness results of PDF and Offline algorithms over 3 runs to ease comparison. In these tables, we highlight a result if the difference in performance is greater than the standard deviation of both PDF and Offline network results. Although PDF achieves better results than Offline, average over 3 runs brings the results of the two algorithms close to each other in most cases. However, in Table \ref{table:mean_and_std_gen} we see that PDF performs better than Offline in 8 cases. In robustness results presented in Table \ref{table:mean_and_std_robust} PDF performs on par with Offline in most cases and better in 1 case. Overall, PDF can still be considered to provide a network with better decoder sharing scheme than Offline.

\begin{table}[h]
\caption{Mean and standard deviation of generalization results of PDF and Offline algorithms over 3 seeds.}
\centering
\resizebox{\columnwidth}{!}{
\begin{tabular}{|c|c|ccccc|ccccc|}
\toprule
 \multicolumn{2}{|c|}{\multirow{2}{*}{Network}} & \multicolumn{ 5}{c|}{$\mathcal{S}$ (mIoU \%) $\uparrow$} & \multicolumn{5}{c|}{$\mathcal{D}$ (RMSE) $\downarrow$} \\ \cmidrule{3-12}
 \multicolumn{2}{|c|}{} & IID & noise & blur & weather & digital & IID & noise & blur & weather & digital \\ \midrule
 \multirow{2}{*}{\rotatebox[origin=c]{90}{CS}}
 & Offline & 73.01{\tiny$\pm0.34$} & 18.02{\tiny$\pm0.21$} & 48.91{\tiny$\pm0.33$} & 35.85{\tiny$\pm0.11$} & 61.13{\tiny$\pm0.48$} & 5.35{\tiny$\pm0.02$} & 14.85{\tiny$\pm0.67$} & 13.25{\tiny$\pm0.11$} & 11.88{\tiny$\pm0.44$} & 7.40{\tiny$\pm0.15$} \\
 & PDF & 72.92{\tiny$\pm0.25$} & \textbf{19.72}{\tiny$\pm1.39$} & 49.19{\tiny$\pm0.22$} & \textbf{36.68}{\tiny$\pm0.29$} & 60.98{\tiny$\pm0.24$} & 5.31{\tiny$\pm0.01$} & 15.09{\tiny$\pm0.52$} & 13.04{\tiny$\pm0.38$} & 12.09{\tiny$\pm0.03$} & 7.52{\tiny$\pm0.33$} \\ 
 \midrule
 \multirow{2}{*}{\rotatebox[origin=c]{90}{NYU}} 
 & Offline & 42.46{\tiny$\pm0.28$} & 4.51{\tiny$\pm0.32$} & 23.87{\tiny$\pm0.32$} & 18.93{\tiny$\pm0.13$} & 26.54{\tiny$\pm0.42$} & 47.98{\tiny$\pm0.24$} & 106.92{\tiny$\pm3.88$} & \textbf{81.66}{\tiny$\pm0.82$} & 80.88{\tiny$\pm1.18$} & 66.40{\tiny$\pm0.53$} \\
 & PDF & 42.28{\tiny$\pm0.19$} & \textbf{5.04}{\tiny$\pm0.33$} & \textbf{24.27}{\tiny$\pm0.22$} & 18.83{\tiny$\pm0.14$} & \textbf{27.14}{\tiny$\pm0.08$} & \textbf{47.22}{\tiny$\pm0.49$} & 113.53{\tiny$\pm10.19$} & 83.89{\tiny$\pm0.70$} & \textbf{78.14}{\tiny$\pm1.39$} & \textbf{65.23}{\tiny$\pm0.85$} \\ 
 \bottomrule
\end{tabular}}
\label{table:mean_and_std_gen}
\end{table}

\begin{table}[h]
\caption{Mean and standard deviation of robustness of PDF and Offline algorithms results over 3 seeds.}
\centering
\resizebox{\columnwidth}{!}{
\begin{tabular}{|c|cc|cc|cc|cc|}
\toprule
\multirow{3}{*}{Network} & \multicolumn{4}{c|}{CS} & \multicolumn{4}{c|}{NYU} \\ \cmidrule{2-9}
 & \multicolumn{2}{c|}{$\mathcal{S}$ (mIoU \%) $\uparrow$} & \multicolumn{2}{c|}{$\mathcal{D}$ (RMSE) $\downarrow$} & \multicolumn{ 2}{c|}{$\mathcal{S}$ (mIoU \%) $\uparrow$} & \multicolumn{2}{c|}{$\mathcal{D}$ (RMSE) $\downarrow$} \\ \cmidrule{2-9}
 & low $\epsilon$ & high $\epsilon$ & low $\epsilon$ & high $\epsilon$ & low $\epsilon$ & high $\epsilon$ & low $\epsilon$ & high $\epsilon$ \\ \cmidrule{1-9}
Offline & 46.02{\tiny$\pm0.27$} & 8.12{\tiny$\pm0.33$} & 15.27{\tiny$\pm0.13$} & 44.90{\tiny$\pm0.46$} & 20.46{\tiny$\pm0.40$} & 3.55{\tiny$\pm0.29$} & 110.72{\tiny$\pm1.19$} & 308.61{\tiny$\pm4.96$} \\ 
PDF & 45.75{\tiny$\pm0.34$} & 8.07{\tiny$\pm0.47$} & 15.19{\tiny$\pm0.13$} & 44.93{\tiny$\pm1.03$} & 20.38{\tiny$\pm0.24$} & 3.58{\tiny$\pm0.06$} & \textbf{107.71}{\tiny$\pm0.19$} & 304.94{\tiny$\pm4.74$} \\ 
\bottomrule
\end{tabular}}
\label{table:mean_and_std_robust}
\end{table}

\begin{table}[h]
\caption{Mean of generalization results of networks obtained at various stages of the PDF algorithm over 3 seeds.}
\centering
\resizebox{\columnwidth}{!}{
\begin{tabular}{|c|c|ccccc|ccccc|}
\toprule
 \multicolumn{2}{|c|}{\multirow{2}{*}{Network}} & \multicolumn{ 5}{c|}{$\mathcal{S}$ (mIoU \%) $\downarrow$} & \multicolumn{5}{c|}{$\mathcal{D}$ (RMSE) $\uparrow$} \\ \cmidrule{3-12}
 \multicolumn{2}{|c|}{} & IID & noise & blur & weather & digital & IID & noise & blur & weather & digital \\ \midrule
 \multirow{5}{*}{\rotatebox[origin=c]{90}{CS}} 
 & PDF-5 & 72.92 & 19.72 & 49.19 & 36.68 & 60.98 & 5.31 & 15.09 & 13.04 & 12.09 & 7.52 \\ 
 & PDF-4 & 72.41 & 18.67 & 48.77 & 35.54 & 60.93 & 5.33 & 15.70 & 13.15 & 11.89 & 7.41 \\ 
 & PDF-3 & 72.63 & 19.25 & 48.88 & 35.86 & 60.69 & 5.33 & 14.25 & 13.11 & 11.98 & 7.57 \\ 
 & PDF-2 & 72.99 & 16.39 & 49.08 & 36.37 & 61.13 & 5.32 & 15.78 & 12.94 & 12.21 & 7.36 \\ 
 & PDF-1 & 72.17 & 19.27 & 48.34 & 35.72 & 60.76 & 5.34 & 15.11 & 13.20 & 12.29 & 7.34 \\ 
 \midrule
 \multirow{5}{*}{\rotatebox[origin=c]{90}{NYU}} 
 & PDF-5 & 42.28 & 5.04 & 24.27 & 18.83 & 27.14 & 47.22 & 113.53 & 83.89 & 78.14 & 65.23 \\ 
 & PDF-4 & 42.44 & 4.64 & 24.10 & 19.04 & 26.79 & 47.84 & 115.47 & 83.13 & 80.34 & 65.34 \\ 
 & PDF-3 & 42.12 & 4.84 & 23.88 & 18.73 & 26.28 & 47.84 & 99.48 & 83.44 & 82.21 & 64.92 \\ 
 & PDF-2 & 41.90 & 5.09 & 23.95 & 18.95 & 26.48 & 47.70 & 120.09 & 83.62 & 81.66 & 65.34 \\ 
 & PDF-1 & 42.33 & 4.30 & 24.00 & 18.80 & 26.76 & 48.00 & 120.28 & 83.79 & 80.40 & 65.53 \\ 
 \bottomrule
\end{tabular}}
\label{table:mean_gen}
\end{table}

\begin{table}[h]
\caption{Mean of robustness results of networks obtained at various stages of the PDF algorithm over 3 seeds.}
\centering
\begin{tabular}{|c|cc|cc|cc|cc|}
\toprule
\multirow{3}{*}{Network} & \multicolumn{4}{c|}{CS} & \multicolumn{4}{c|}{NYU} \\ \cmidrule{2-9}
 & \multicolumn{2}{c|}{$\mathcal{S}$ (mIoU \%) $\uparrow$} & \multicolumn{2}{c|}{$\mathcal{D}$ (RMSE) $\downarrow$} & \multicolumn{ 2}{c|}{$\mathcal{S}$ (mIoU \%) $\uparrow$} & \multicolumn{2}{c|}{$\mathcal{D}$ (RMSE) $\downarrow$} \\ \cmidrule{2-9}
 & low $\epsilon$ & high $\epsilon$ & low $\epsilon$ & high $\epsilon$ & low $\epsilon$ & high $\epsilon$ & low $\epsilon$ & high $\epsilon$ \\ \cmidrule{1-9}
PDF-5 & 45.75 & 8.07 & 15.19 & 44.93 & 20.38 & 3.58 & 107.71 & 304.94 \\ 
PDF-4 & 46.21 & 8.08 & 15.22 & 44.85 & 20.29 & 3.48 & 110.98 & 309.45 \\ 
PDF-3 & 46.13 & 8.14 & 15.28 & 44.72 & 20.00 & 3.34 & 110.38 & 313.14 \\ 
PDF-2 & 46.44 & 8.64 & 15.16 & 43.57 & 20.33 & 3.39 & 110.74 & 311.96 \\ 
PDF-1 & 45.46 & 8.40 & 15.26 & 45.60 & 20.22 & 3.33 & 109.66 & 300.99 \\ 
\bottomrule
\end{tabular}
\label{table:mean_robust}
\end{table}

\begin{table}[h]
\caption{Standard deviation of generalization results of different networks over 3 seeds.}
\centering
\begin{tabular}{|c|c|ccccc|ccccc|}
\toprule
 \multicolumn{2}{|c|}{\multirow{2}{*}{Network}} & \multicolumn{ 5}{c|}{$\mathcal{S}$ (mIoU \%)} & \multicolumn{5}{c|}{$\mathcal{D}$ (RMSE)} \\ \cmidrule{3-12}
 \multicolumn{2}{|c|}{} & IID & noise & blur & weather & digital & IID & noise & blur & weather & digital \\ \midrule
 \multirow{5}{*}{\rotatebox[origin=c]{90}{CS}} & One-De & 0.25 & 0.54 & 0.09 & 1.31 & 0.24 & 0.01 & 0.10 & 0.19 & 0.50 & 0.28 \\ 
 & Sep-De & 0.09 & 1.76 & 0.22 & 0.92 & 0.06 & 0.01 & 1.10 & 0.15 & 0.23 & 0.05 \\
 & Offline & 0.34 & 0.21 & 0.33 & 0.11 & 0.48 & 0.02 & 0.67 & 0.11 & 0.44 & 0.15 \\ 
 & PDF-5 & 0.25 & 1.39 & 0.22 & 0.29 & 0.24 & 0.01 & 0.52 & 0.38 & 0.03 & 0.33 \\ 
 & PDF-4 & 0.57 & 0.38 & 0.14 & 0.44 & 0.10 & 0.01 & 0.42 & 0.20 & 0.40 & 0.16 \\ 
 & PDF-3 & 0.45 & 0.74 & 0.09 & 0.07 & 0.17 & 0.02 & 1.06 & 0.11 & 0.03 & 0.33 \\ 
 & PDF-2 & 0.39 & 0.76 & 0.35 & 0.56 & 0.76 & 0.02 & 0.78 & 0.17 & 0.38 & 0.15 \\ 
 & PDF-1 & 0.63 & 1.01 & 0.09 & 0.77 & 0.40 & 0.03 & 2.14 & 0.17 & 0.21 & 0.12 \\ 
 & PDF-5-No-$\mathcal{A}$ & 0.38 & 0.88 & 0.20 & 0.93 & 0.59 & 0.01 & 1.15 & 0.16 & 0.32 & 0.04 \\ \midrule
 \multirow{5}{*}{\rotatebox[origin=c]{90}{NYU}} & One-De & 0.42 & 0.67 & 0.48 & 0.32 & 0.20 & 0.10 & 11.35 & 0.49 & 1.17 & 0.37 \\ 
 & Sep-De & 0.40 & 0.10 & 0.15 & 0.33 & 0.25 & 0.16 & 5.43 & 0.77 & 0.77 & 0.96 \\ 
 & Offline & 0.28 & 0.32 & 0.32 & 0.13 & 0.42 & 0.24 & 3.88 & 0.82 & 1.18 & 0.53 \\ 
 & PDF-5 & 0.19 & 0.33 & 0.22 & 0.14 & 0.08 & 0.49 & 10.19 & 0.70 & 1.30 & 0.85 \\ 
 & PDF-4 & 0.41 & 0.48 & 0.45 & 0.11 & 0.29 & 0.27 & 8.15 & 1.01 & 1.14 & 1.74 \\ 
 & PDF-3 & 0.23 & 0.79 & 0.33 & 0.19 & 0.41 & 0.14 & 7.70 & 1.33 & 2.54 & 0.64 \\ 
 & PDF-2 & 0.68 & 0.26 & 0.55 & 0.27 & 0.41 & 0.16 & 9.95 & 1.01 & 2.45 & 1.10 \\ 
 & PDF-1 & 0.35 & 0.52 & 0.51 & 0.23 & 0.50 & 0.12 & 4.24 & 0.93 & 2.42 & 0.71 \\ 
 & PDF-5-No-$\mathcal{A}$ & 0.10 & 1.06 & 0.29 & 0.09 & 0.22 & 0.24 & 4.07 & 0.77 & 0.40 & 0.71 \\ \bottomrule
\end{tabular}
\label{table:std_gen}
\end{table}

\begin{table}[h]
\caption{Standard deviation of robustness results of different networks over 3 seeds.}
\centering
\begin{tabular}{|c|cc|cc|cc|cc|}
\toprule
\multirow{3}{*}{Network} & \multicolumn{4}{c|}{CS} & \multicolumn{4}{c|}{NYU} \\ \cmidrule{2-9}
 & \multicolumn{2}{c|}{$\mathcal{S}$ (mIoU \%)} & \multicolumn{2}{c|}{$\mathcal{D}$ (RMSE)} & \multicolumn{ 2}{c|}{$\mathcal{S}$ (mIoU \%)} & \multicolumn{2}{c|}{$\mathcal{D}$ (RMSE)} \\ \cmidrule{2-9}
 & low $\epsilon$ & high $\epsilon$ & low $\epsilon$ & high $\epsilon$ & low $\epsilon$ & high $\epsilon$ & low $\epsilon$ & high $\epsilon$ \\ \cmidrule{1-9}
One-De & 0.64 & 0.60 & 0.11 & 0.33 & 0.28 & 0.06 & 1.09 & 2.12 \\ 
Sep-De & 0.73 & 0.61 & 0.06 & 1.03 & 0.14 & 0.16 & 0.19 & 3.55 \\ 
Offline & 0.27 & 0.33 & 0.13 & 0.46 & 0.40 & 0.29 & 1.14 & 4.96 \\ 
PDF-5 & 0.34 & 0.47 & 0.13 & 1.03 & 0.24 & 0.06 & 0.19 & 4.74 \\ 
PDF-4 & 0.78 & 0.29 & 0.09 & 0.74 & 0.18 & 0.19 & 0.50 & 8.77 \\ 
PDF-3 & 0.27 & 0.08 & 0.10 & 1.08 & 0.35 & 0.05 & 0.55 & 2.96 \\ 
PDF-2 & 0.35 & 0.50 & 0.09 & 0.54 & 0.25 & 0.10 & 0.79 & 4.42 \\ 
PDF-1 & 0.84 & 0.38 & 0.06 & 1.00 & 0.38 & 0.14 & 0.20 & 3.05 \\ 
PDF-5-No-$\mathcal{A}$ & 0.05 & 0.03 & 0.20 & 0.36 & 0.22 & 0.17 & 15.53 & 25.76 \\ \bottomrule
\end{tabular}
\label{table:std_robust}
\end{table}

\end{document}